\documentclass{article}

\usepackage{arxiv}

\usepackage[utf8]{inputenc} 
\usepackage[T1]{fontenc}    
\usepackage{hyperref}       
\usepackage{url}            
\usepackage{booktabs}       
\usepackage{amsfonts}       
\usepackage{nicefrac}       
\usepackage{microtype}      
\usepackage{lipsum}
\usepackage{graphicx}
\graphicspath{ {./images/} }
\usepackage{amsmath}
\usepackage{amssymb}
\usepackage{multirow}
\usepackage{booktabs} 
\usepackage{tabularx}
\usepackage{adjustbox}
\usepackage{float}
\usepackage{svg}
\usepackage{graphicx,verbatim}
\usepackage{soul}
\usepackage{ragged2e}
\usepackage[table]{xcolor}
\usepackage{soul}
\definecolor{oraclegray}{RGB}{230,230,230} 
\usepackage{pifont}
\usepackage{amssymb}
\newcommand{\cmark}{\ding{51}} 
\newcommand{\xmark}{\ding{55}} 

\title{Uncertainty-Aware Information Pursuit for Interpretable and Reliable Medical Image Analysis}

\author{
 Md Nahiduzzaman \\
  School of Computing Technologies\\
  RMIT University\\
  Melbourne VIC 3000, Australia \\
  \texttt{s4045807@student.rmit.edu.au} \\
  \And
 Steven Korevaar \\
  School of Computing Technologies\\
  RMIT University\\
  Melbourne VIC 3000, Australia \\
  \texttt{steven.korevaar@rmit.edu.au} \\
  \And
  Zongyuan Ge \\
  Faculty of Engineering\\
  Monash University\\
  Clayton VIC 3800, Australia \\
  \texttt{zongyuan.ge@monash.edu} \\
  \And
  Feng Xia \\
  School of Computing Technologies\\
  RMIT University\\
  Melbourne VIC 3000, Australia \\
  \texttt{feng.xia@rmit.edu.au} \\
  \And
  Alireza Bab-Hadiashar \\
  School of Engineering\\
  RMIT University\\
  Melbourne VIC 3000, Australia \\
  \texttt{alireza.bab-hadiashar@rmit.edu.au} \\
     \And
 Ruwan Tennakoon \\
  School of Computing Technologies\\
  RMIT University\\
  Melbourne VIC 3000, Australia \\
  \texttt{ruwan.tennakoon@rmit.edu.au} \\
}

\begin{document}
\maketitle
\begin{abstract}
To be adopted in safety-critical domains like medical image analysis, AI systems must provide human-interpretable decisions. Variational Information Pursuit (V-IP) offers an interpretable-by-design framework by sequentially querying input images for human-understandable concepts, using their presence or absence to make predictions. However, existing V-IP methods overlook sample-specific uncertainty in concept predictions, which can arise from ambiguous features or model limitations, leading to suboptimal query selection and reduced robustness. 
In this paper, we propose an interpretable and uncertainty-aware framework for medical imaging that addresses these limitations by accounting for upstream uncertainties in concept-based, interpretable-by-design models. Specifically, we introduce two uncertainty-aware models, EUAV-IP and IUAV-IP, that integrate uncertainty estimates into the V-IP querying process to prioritize more reliable concepts per sample. EUAV-IP skips uncertain concepts via masking, while IUAV-IP incorporates uncertainty into query selection implicitly for more informed and clinically aligned decisions. Our approach allows models to make reliable decisions based on a subset of concepts tailored to each individual sample, without human intervention, while maintaining overall interpretability.
We evaluate our methods on five medical imaging datasets across four modalities: dermoscopy, X-ray, ultrasound, and blood cell imaging. The proposed IUAV-IP model achieves state-of-the-art accuracy among interpretable-by-design approaches on four of the five datasets, and generates more concise explanations by selecting fewer yet more informative concepts. These advances enable more reliable and clinically meaningful outcomes, enhancing model trustworthiness and supporting safer AI deployment in healthcare.  Our code and models are available
at: \url{https://github.com/Nahiduzzaman09/UAV-IP}.
\end{abstract}


\section{Introduction}
\label{sec:introduction}
Given the growing prevalence of deep learning (DL) in safety-sensitive domains like medical diagnosis \cite{liu2023semi, guo20243d, gong2025boundary}, ensuring model interpretability is crucial for cultivating trust and facilitating responsible decision-making. 
Despite their impressive performance, DL models often function a ``black box'' \cite{castelvecchi2016can, shwartz2022opening, lipton2018mythos}, making it difficult for users to fully trust their outputs, or use them in collaborative settings. To address this challenge, a substantial body of research has focused on developing methods to explain how DL models make decisions, particularly through post-hoc (PH) analysis applied after training. A key advantage of PH methods is that they preserve the original performance. However, PH explanations often fall short of being truly meaningful or faithful to the model’s internal reasoning \cite{ribeiro2016should, adebayo2018sanity, rudin2019stop, geirhos2024don}. As a result, interpretable-by-design (IBD) models have recently gained significant attention in the research community. These IBD models are designed such that the final model is fully human-interpretable, decision making processes are transparent and understandable while maintaining strong performance.

One of the most prominent IBD models in medical applications is the Concept Bottleneck Model (CBM) \cite{koh2020concept}. CBMs work by mapping the extracted features from the input image to an intermediate set of human-understandable concepts such as ``blue-whitish veil'' or ``regression structures'' in the context of skin cancer diagnosis. These concepts are then passed through a single linear layer to produce the final prediction. The use of this intermediate concept space offers a transparent explanation of the model’s reasoning, as the concepts are directly aligned with clinical knowledge. 
However, to maintain interpretability, CBMs \cite{koh2020concept, kim2023probabilistic, gao2024evidential} employ linear classifiers as their final prediction layers, which can be limiting in domains such as medical imaging, where diagnostic features often interact in complex, non-linear ways \cite{chattopadhyay2022interpretable}. Furthermore, as discussed in Section \ref{sec:CBM}, CBMs are susceptible to information leakage and rely on a fixed set of concepts for decision-making, even when some concepts may be irrelevant or unreliable for specific data points.

The Variational Information Pursuit (V-IP) framework \cite{chattopadhyay2022interpretable, chattopadhyayvariational, chattopadhyay2024bootstrapping, kolek2025learning, ge2025ip}, which emulates the human decision-making process, addresses some of the above limitations. Rather than processing all information at once or combining concepts linearly, V-IP observes parts of the input sequentially, selecting each based on its expected informativeness. In skin cancer example, it poses clinically meaningful queries, such as ``Is a blue-whitish veil present?'' or ``Are there regression structures?,'' sequentially, and uses the accumulated history of query-answer pairs to decide what concept to examine next. This iterative process allows the model to refine its prediction step-by-step and make decisions only based on a subset of concepts that are more relevant for a specific input without any human intervention.

\emph{A key limitation of the original V-IP framework lies in its inability to dynamically adapt to upstream uncertainty that arises at the level of individual test inputs.}\footnote{Here, upstream uncertainty primarily refers to uncertainty in the concept predictor (CP), i.e., the bottleneck model used by V-IP to generate query answers from raw inputs.}
While V-IP learns query policies that are optimal on average under the training distribution, its query selection is governed by a globally learned model of concept–label relationships and does not explicitly account for uncertainty that is \textit{instance-specific at test time}. In particular, the framework assumes that concept predictions provided by the CP are reliable, even when the visual evidence supporting a concept is weak or absent for a given input.

For example, in bowel cancer prediction, “abnormal lymph node appearance” may be a highly predictive concept in the training data, and V-IP may therefore learn to prioritise querying it. However, for a specific test case in which lymph nodes are not visible in the patient’s imaging data, this concept becomes inherently unreliable. Despite this, the CP, often implemented as a deterministic deep neural network, still produces a binary yes/no answer, which is treated as equally informative as in cases where the concept is clearly observable. As a result, the corresponding question–answer pair provides little diagnostic value, yet V-IP lacks a mechanism to downweight or bypass such queries based on instance-specific uncertainty.

In this paper, we address the previously identified limitation of the V-IP framework and investigate how instance-level uncertainties can be effectively integrated into the V-IP framework.
In our approach, we use instance-level uncertainty estimates for each concept to guide the ``next-best'' query selection process and to update V-IP's belief about a given class.
We explore two key strategies for integrating uncertainty:
1) {Explicit Uncertainty Handling} via \emph{Explicit Uncertainty-Aware V-IP (EUAV-IP)}: Here, highly uncertain concept predictions are excluded through a masking mechanism. 
2) {Implicit Uncertainty Integration} via \emph{Implicit Uncertainty-Aware V-IP (IUAV-IP)}: Rather than masking uncertain queries, IUAV-IP feeds both the concept prediction and its associated uncertainty into the querier. This enables the model to reason about uncertainty directly, instead of ignoring it. 

Our approach is agnostic about the uncertainty quantification method used. It can leverage many established methods to quantify upstream uncertainties, allowing it to handle both \emph{aleatoric} and \emph{epistemic} uncertainties. Furthermore, uncertainty aware V-IP models (UAV-IP) can provide a classification based on a subset of more certain concepts and ignore the uncertain ones without any expert intervention, unlike CBMs \cite{gao2024evidential, kim2023probabilistic}.
Together, these contributions enable UAV-IP to deliver more efficient, flexible, and clinically realistic interpretations. Importantly, our goal is not merely to maximize predictive accuracy, but to ensure that model decisions remain transparent, verifiable, and clinically meaningful, even in regimes where standard performance metrics may appear saturated.
The main contributions of this work are summarized as follows:

\begin{itemize}
    \item To the best of our knowledge, this is the first work to incorporate instance-level uncertainty into the V-IP framework, enabling models to make decisions based on a \textbf{more reliable subset of concepts per instance}, without requiring human intervention, while maintaining interpretability.
    
    \item In addition to the explicit EUAV-IP approach that skips uncertain queries, we propose an implicit integration of uncertainty into the V-IP framework, wherein the querier is modified to consider both the \emph{query history} (i.e., previously asked queries and their answers) and the \emph{uncertainty} of all candidate queries when selecting the next one. Furthermore, we introduce a \textbf{penalty-based loss function} that encourages the model to prioritize high-certainty, informative concepts early in the reasoning process. Finally, we incorporate a \textbf{hybrid stopping criterion} that adaptively determines when to terminate the querying process, thereby reducing the number of queries while maintaining or improving predictive performance.

    \item Our results on five public medical imaging datasets with four different modalities demonstrate that UAV-IPs generate more concise and meaningful explanations while maintaining high performance compared to state-of-the-art methods.
\end{itemize}
\section{Related Work \& Background}
\label{PR}

\subsection{Interpretable-by-Design (IBD) Methods}

\subsubsection{Prototype-Based Models}
Prototype-based models learn parts of an image that resemble human-understandable patterns. ProtoPNet~\cite{chen2019looks} compares image patches to learned prototypes for each class, offering intuitive visual explanations. However, it needs many prototypes and is slow. ProtoTree~\cite{nauta2021neural} makes this more efficient by organizing prototypes in a tree structure. ProtoAttend~\cite{arik2020protoattend} adds attention to focus on relevant regions. Despite being transparent, these models sometimes learn parts that don't match how humans perceive a concept ~\cite{wickramanayake2021comprehensible} and defining prototypes may be irrelevant in some medical imaging tasks.

\subsubsection{Concept Bottleneck Models (CBMs)}
\label{sec:CBM}
CBMs predict a set of human-labeled concepts before making the final decision. Koh et al.~\cite{koh2020concept} introduced this idea by adding a concept layer between the input and output. Changes to these concept values directly affect the prediction, offering clear insight into the model’s reasoning. Later works improved this by ensuring unique concept mappings~\cite{wickramanayake2021comprehensible}, focusing on relevant image regions~\cite{patricio2023coherent}, and adding decoders to reconstruct the input image~\cite{sarkar2022framework}.
Some works aimed to reduce labeling costs. Yuksekgonul et al.~\cite{yuksekgonul2022post} proposed post-hoc CBMs that only train the final layer using concept activation vectors. Others~\cite{oikarinen2023label, yang2023language} used Generative Pretrained Transformer (GPT) and Contrastive Language–Image Pretraining (CLIP) to generate concepts automatically.

Despite recent advances, many CBMs continue to suffer from limited expressivity due to their reliance on linear mappings from concepts to predictions~\cite{koh2020concept}, which may be inadequate for modeling complex decision boundaries~\cite{janssen1997compositionality}. Additionally, CBMs are vulnerable to \emph{information leakage}\cite{mahinpei2021promises}, where latent features inadvertently carry spurious or non-interpretable signals into the prediction layer. This undermines their goal of human-understandable reasoning. Some works attempt to address these issues, for example, logic-based approaches like concept rule learning \cite{gao2025learning} enforce discrete boolean operations between concepts and labels, offering stronger safeguards against information leakage and enforcing fully-interpretable classification.

\subsubsection{Active Testing-Based Models}
To move beyond the linear additivity concern in CBMs, Chattopadhyay et al.~\cite{chattopadhyay2022interpretable} proposed the Information Persuit (IP), which selects only the most informative queries to classify an input. This follows the idea of active testing~\cite{geman1996active}, similar to how a doctor might ask follow-up questions before making a diagnosis. However, IP relies on complex generative models, making it hard to scale.
Later, V-IP model~\cite{chattopadhyayvariational} simplified this by jointly learning the query selection (querier) and classifier using cross-entropy loss. 

Chattopadhyay et al.~\cite{chattopadhyay2024bootstrapping} advanced this direction by bootstrapping V-IP with GPT and CLIP to automatically generate and answer semantic queries, removing the need for manual concept annotations. Kolek et al.~\cite{kolek2025learning} made the V-IP framework more scalable by learning a task-sufficient and interpretable query dictionary directly from CLIP embeddings, replacing handcrafted concepts with data-driven ones. Their alternating optimization jointly updates the query dictionary and the V-IP network to improve the sufficiency and diversity of queries.  Our work targets a complementary axis: we model instance-specific uncertainty at test time to adapt which query to ask next and how much to trust its answer, and we introduce uncertainty-aware stopping.
Thus, the above methods improve the process of obtaining or curating queries on average, whereas our method enhances the selection and utilization of queries for each image.
In principle, our uncertainty-aware policy can operate on top of a learned or bootstrapped dictionary.

\subsection{Uncertainty in IBD Models}

Uncertainty modeling is widely used to improve the interpretability, robustness, and trustworthiness of DL models~\cite{gal2016dropout, sale2024label}. In general, uncertainty in machine learning is categorized into two types: \emph{epistemic uncertainty} (also known as model uncertainty), which arises from limited training data or model capacity, and \emph{aleatoric uncertainty} (data uncertainty), which stems from inherent noise or ambiguity in the input data~\cite{hullermeier2021aleatoric}.

Kim et al.~\cite{kim2023probabilistic} introduced Probabilistic Concept Bottleneck Model (ProbCBM), an extension of traditional CBMs in which each concept is modeled as a distribution rather than a single deterministic output. This probabilistic formulation captures the uncertainty in concept predictions and propagates it to the final classifier, leading to more interpretable outputs. In contrast,  Evidential Concept Embedding Model (evi-CEM) by Gao et al. \cite{gao2024evidential} models uncertainty with the goal of improving concept prediction quality, particularly in weakly supervised settings, by identifying and correcting misaligned concepts using a small set of labeled examples and Concept Activation Vectors (CAVs).

Despite these strengths, these models inherit a key limitation from standard CBMs: the linear additivity constraint and the issue of information leakage~\cite{mahinpei2021promises, gao2025learning}. Additionally, these methods require all concepts to make a decision. When some concepts are uncertain, they depend on human experts to review and adjust them \cite{gao2024evidential}, or substitute them with class averages \cite{kim2023probabilistic}, which overlooks sample-specific nuances.
\section{Methodology}
We consider the problem of \emph{supervised query-based classification}, where predictions are made via responses to a set of human-interpretable binary queries (or concepts). Let the random variable \( X \in \mathcal{X} \) denote an input (e.g., a medical image) and \( Y \in \mathcal{Y} \) its associated class label. We assume access to a predefined set of \( M \) semantic queries
\(
\mathcal{Q} = \{ q_1, q_2, \dots, q_M \},
\)
typically curated by domain experts.\footnote{Large language models (LLMs) may also be used to automatically derive concept-based queries~\cite{radford2021learning, oikarinen2023label, yang2023language, chattopadhyay2024bootstrapping}.}
Each query corresponds to an underlying (latent) binary concept function
\(
q_m : \mathcal{X} \to \{-1, +1\},
\)
where \( q_m(X) = +1 \) indicates the presence of the concept and \( -1 \) its absence.  
For example, if \( X = x \) is a dermoscopic image and \( q_m \) corresponds to the question ``Is the lesion palpable/elevated?'', then \( q_m(x) = +1 \) when the feature is present.

In practice, the true concepts \( \{ q_m \} \) are not available for test instances. Instead, their values are estimated using a \emph{concept predictor (CP)},
\[
q_\theta : \mathcal{X} \to \{-1, +1\}^M,
\]
parameterized by \( \theta \), which outputs a vector of predicted concept responses
\(
\hat{\mathbf{q}}(X) = (\hat{q}_1(X), \dots, \hat{q}_M(X)).
\)
The CP may be instantiated using any suitable model class, including deep neural networks.

We adopt the standard modeling assumption used in concept-based and V-IP frameworks that the set of concepts is \emph{label-sufficient}, i.e.,
for all \( (x,y) \in (\mathcal{X}, \mathcal{Y}) \), the following holds:  
    $P(y \mid x) = P(y \mid x' \in \mathcal{X}: \{ q_m(x') = q_m(x)\}_{m=1}^M)$.

We follow prior work~\cite{chattopadhyayvariational} and consider a \emph{sequential querying strategy}. At each step, the model selects a query, observes its predicted response, and updates its belief over the class label.

\subsection{Preliminaries: Information Pursuit}

In Information Pursuit (IP), the authors formulate efficient query composition for prediction as the following optimization problem~\cite{chattopadhyay2022interpretable}:
\[
\begin{aligned}
\min_{\pi} \quad & \mathbb{E}_X \big[ |\mathcal{H}^\pi_{1:t^\pi(X)}(X)| \big] \\
\text{s.t.} \quad &
\mathbb{E}_X \Big[ d\big( p(Y\mid X), p(Y \mid \mathcal{H}^\pi_{1:t^\pi(X)}(X)) \big) \Big] \le \delta.
\end{aligned}
\]
Here, \(
\mathcal{H}_{1:k}^\pi(X)
=
\big[ (q_{i_l}, q_{i_l}(X)) \big]_{l=1}^k
\in \mathbb{H},
\) 
denotes the ordered history of query--answer pairs generated for an input $X$ by a policy $\pi$.
At step $k$, the query index $i_k$ is selected according to
\[
q_{i_k} = \pi\!\left( \mathcal{H}^\pi_{1:k-1}(X) \right).
\]
The random variable $t^\pi(X)$ denotes the number of queries selected by the policy $\pi$ before it terminates,
$d(\cdot,\cdot)$ is a divergence measure, and $\delta$ is an approximation constant. When $d$ is chosen as KL-divergence, the constraint admits an information-theoretic interpretation~\cite{chattopadhyay2022interpretable}:
\[
\begin{aligned}
\min_{\pi} \quad & \mathbb{E}_X \big[ |\mathcal{H}^\pi_{1:t^\pi(X)}(X)| \big] \\
\text{s.t.} \quad &
\sum_{k=1}^{\tau_\pi}
I\big(Y; S_k^\pi(X) \mid S_{k-1}^\pi(X)\big)
\ge I(X;Y) - \delta.
\end{aligned}
\]
where 
\(
S_k^\pi(x_{\mathrm{obs}})
\;\triangleq\;
\Big\{
x \in \mathcal{X} \;\Big|\; q_j(x) = q_j(x_{\mathrm{obs}}), \;\forall j=1,\dots,k
\Big\}
\) contain a set of data points from the training dataset where query--answers match that of the observed data produced by \(\pi\), $\tau^\pi = \max \{ t^\pi(x): x \in \mathcal{X}\}$, and \(I(\cdot;\cdot)\) is the mutual information.
Solving the above problem exactly is intractable.
The authors propose a greedy approximation where
the first query is selected as
\[
q_{i_1} = \arg\max_{q \in \mathcal{Q}} I(q(X); Y),
\]
and subsequent queries are chosen adaptively:
\[
q_{i_{k+1}}
=
\arg\max_{q \in \mathcal{Q}}
I\big(q(X); Y \mid S_k(x_{\mathrm{obs}})\big).
\]
The procedure stops when the maximum conditional mutual information falls below a threshold.
Information Pursuit (IP) relies on deep generative models to estimate the joint distribution between $q(X)$ and $Y$, 
which is required to compute the mutual information terms in the above equations, making the framework difficult to scale.
Building on the insight that the next best query $q_{k+1}$ is the query $q^\ast$ whose answer minimizes the 
Kullback--Leibler (KL) divergence between the true posterior $P(Y \mid X)$ and the updated posterior
$P\!\left( Y \mid q^\ast(X), \mathcal{H}_{1:k}^\pi(X) \right)$,
Variational Information Pursuit (V-IP) proposes a scalable variational objective given by~\cite{chattopadhyayvariational}:
\begin{equation}
\mathcal{L}_{\mathrm{VIP}}
=
\mathbb{E}_{X, \mathcal{H}}
\left[
D_{\mathrm{KL}}\!\left(
P\!\left( Y \mid X \right)
\;\big\|\;
P_{\eta}\!\left( Y \mid q_{\gamma}(X), \mathcal{H} \right)
\right)
\right].
\label{eqn:VIP_loss}
\end{equation}

Here, $\mathcal{H} \in \mathbb{H}$ denotes a random variable drawn from the set of all possible finite-length query--answer histories, 
$q_{\gamma} = \pi_{\gamma}(\mathcal{H})$, and
\(
P_{\eta}\!\left( Y \mid q_{\gamma}(X), \mathcal{H} \right)
:= f_{\eta}\!\left( \{ q_{\gamma}  \cup \mathcal{H} \} \right)
\). The expectation above is jointly over the data distribution $P(X)$ and the distribution of histories $P(\mathcal{H} \mid X)$.
The two learned components of this framework are:
\begin{enumerate}
    \item \textbf{Querier} \(
    \pi_\gamma : \mathbb{H} \to \mathcal{Q}
    \), a policy that selects the next query based on the accumulated history.
    
    \item \textbf{Probabilistic classifier} \(
    f_\eta : \mathbb{H} \to \Delta(\mathcal{Y})
    \), which predicts a distribution over class labels given the current history.
\end{enumerate}

The querying process continues until a termination criterion \( t^\pi(x) \) is met, yielding a final history
\( \mathcal{H}_{1:t^\pi(x)}^\pi(x) \).
This history serves both as the basis for prediction and as the model’s explanation.

\subsection{Problem Definition}
The existing V-IP framework assumes deterministic query-answer mappings, where each query $q_m(X)$ produces a fixed answer for a given input $X=x$. In practice, queries often exhibit inherent uncertainty due to noisy concept extractors, ambiguous or subjective queries, or stochastic annotation processes, leading to unreliable queries being selected and incorporated into the reasoning process.

To address this, we extend V-IP to stochastic query responses, where each query output is drawn from a conditional distribution \(p(\cdot \mid X, q)\).

For an input \( x \), we denote the uncertainty associated with each predicted concept by
\[
\mathcal{U}_{q_\theta}(x)
=
\big[
\mathcal{U}_{q_1}(x), \dots, \mathcal{U}_{q_M}(x)
\big],
\]
where \( \mathcal{U}_{q_m}(x) \in \mathbb{R}_{\ge 0} \) quantifies the uncertainty of the predicted response \( \hat{q}_m(x) \).
This vector captures \textbf{instance-specific uncertainty} over all concepts.

Our goal is to incorporate these uncertainties into query selection, prediction, and explanation generation, enabling the model to adapt its reasoning to the reliability of each concept at test time. Although uncertainties for all queries can be estimated upfront, a sequential strategy allows the model to update its beliefs dynamically and stop early once the predicted class becomes sufficiently confident.

\begin{figure*}
    \centering
    \includegraphics[width=.9\textwidth]{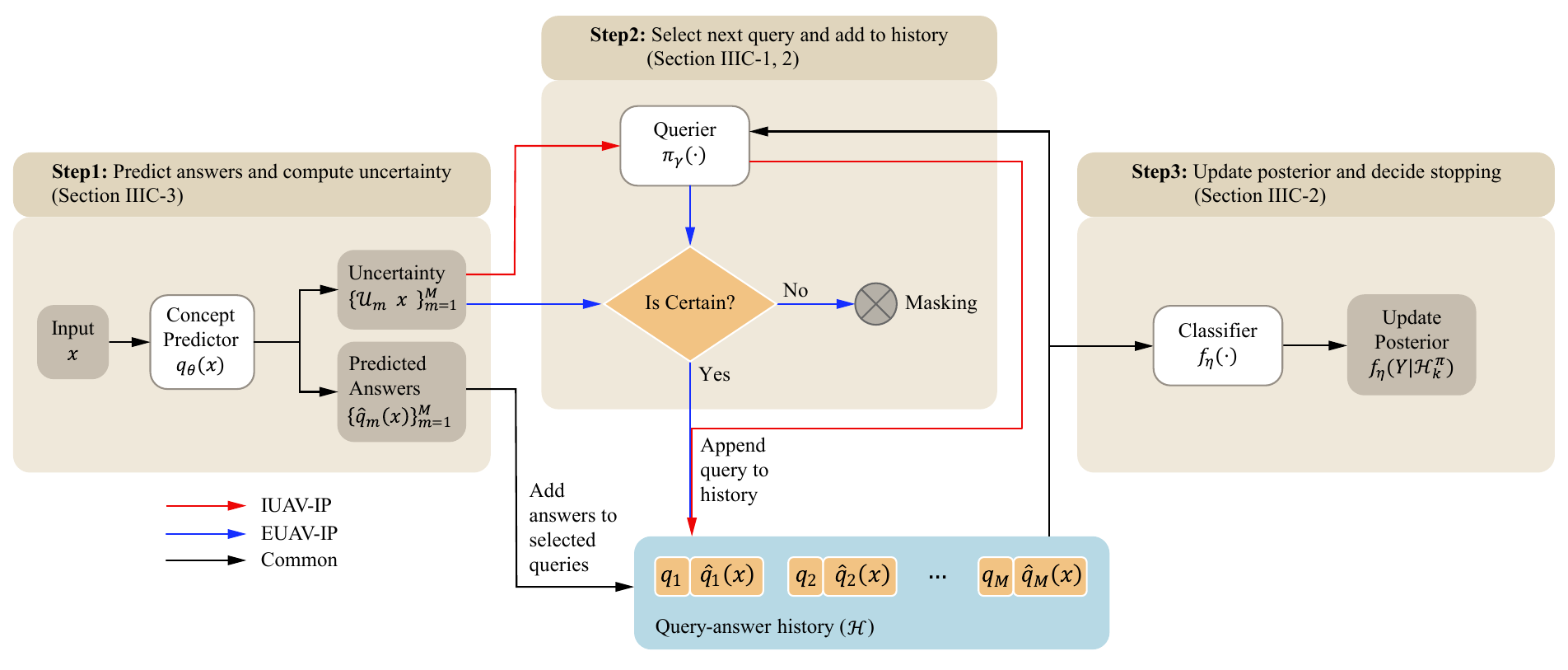}
    \caption{
\textbf{Step-wise pipeline of the Uncertainty-Aware Variational Information Pursuit framework.}
The figure presents the unified inference pipeline, highlighting the distinct information flows for EUAV‑IP (blue arrows) and IUAV‑IP (red arrows). Black arrows indicate components and information flows shared by both variants. During inference, Step 1 is performed once at the beginning, after which Steps 2 and 3 are executed iteratively until the stopping criterion specified in Equation \eqref{equ:stop} is satisfied.
}
\label{fig:framework}
\end{figure*}

\subsection{Proposed Method}


We take inspiration from how human experts navigate uncertainty in clinical decision-making. When faced with ambiguous results from a clinically relevant test, a practitioner may choose to rely instead on other, more reliable sources of information. Following this intuition, we propose a simple yet effective modification to the V-IP framework: the classifier \( f_\eta \) is designed to ignore uncertain concept predictions using a masking strategy. We refer to this approach as \textit{Explicit Uncertainty-Aware Variational Information Pursuit (EUAV-IP)}.

While EUAV-IP introduces a straightforward mechanism to exclude unreliable information during classification, it does not modify the querier’s decision process in selecting the ``next best'' query. Unlike this rigid approach, human decision-makers not only disregard unreliable evidence but also adjust future decisions based on their awareness of uncertainty. Furthermore, experts may still consider ambiguous but potentially valuable tests when no clearly reliable alternatives exist.

To emulate this nuanced reasoning, for the first time, we introduce \textit{Implicit Uncertainty-Aware Variational Information Pursuit (IUAV-IP)}. In IUAV-IP, the querier \( \pi_\gamma \) is conditioned on the estimated uncertainties and learns to prioritize more trustworthy concept signals adaptively. This implicit integration allows the model to reason more flexibly in complex or borderline cases where uncertain concepts may still hold significant predictive value.


An overview of the both UAV-IP architectures, with the modified components highlighted, is shown in Figure~\ref{fig:framework}. The following sections describe our approach to quantifying uncertainty and detail how the query selection and loss functions are modified to support explicit and implicit uncertainty integration.

\subsubsection{Explicit Uncertainty Integration in EUAV-IP}
\label{sec:EUAV-IP}
EUAV-IP requires \emph{instance-level} uncertainty estimatesfor each concept prediction $\mathcal{U}_{q_\theta}(x)$. 
%
These uncertainties are used to define a binary mask \( \Omega(x) \in \{0, 1\}^M \), which indicates whether the corresponding concept prediction should be masked out due to high uncertainty. The mask for each query, \(q_m\), is defined as:
\begin{equation}
    \Omega_m(x) = 
    \begin{cases}
        1 & \text{if } \mathcal{U}_{q_m}(x) \geq \mathcal{T}_{\mathcal{U}} \quad \text{(uncertain)}, \\
        0 & \text{otherwise}.
    \end{cases}
\end{equation}
Here, \( \mathcal{T}_{\mathcal{U}} \) is a predefined or learned threshold that determines whether a concept is considered uncertain. This threshold can be set manually or optimized using a validation set, for example, via a lightweight classifier trained to distinguish between correct and incorrect concept predictions.

The binary uncertainty masks are incorporated into the probabilistic classifier during both training and inference to ignore unreliable concept predictions selectively. The masked classifier is defined as:
\begin{equation}
     f_\eta\left(  \mathcal{H}_{1:k+1}^\pi(x) \odot \Omega(x) \right).
    \label{eqn:UAVIP_classifier}
\end{equation}
The querier remains unchanged. 

\subsubsection{Implicit Uncertainty Integration in IUAV-IP}
\label{sec:IUAV-IP}
While EUAV-IP excludes high-uncertainty concepts using binary masks, the IUAV-IP framework adopts a softer approach. IUAV-IP provides the querier with both the concept history, \(\mathcal{H}_{1:k}\), and its associated uncertainty, \(\mathcal{U}_{q_\theta}(x)\), enabling it to learn a query selection policy that balances informativeness and reliability that is adapted to the current sample.
The modified Querier is defined as:
\[
q_{i_{k+1}} = \pi_\gamma \left ( \mathcal{H}_{1:k}^\pi(x), \mathcal{U}_{q_\theta}(x)\right).
\]
The classifier retains its original form as V-IP and operates exclusively on the history, \(\mathcal{H}^\pi_{1:k+1}\).

\vspace{.5em}
\noindent \textbf{Updated loss function:} While the above model can be directly optimized using the V-IP loss function in \eqref{eqn:VIP_loss}, our ablation study (Sec.~\ref{sec:AS}) revealed that the model does not adequately prioritize certain queries over uncertain ones. To address this and to further encourage the querier to select reliable queries, we introduce an additional penalty term that is applied when a selected concept prediction is incorrect:
\begin{equation}
    \mathcal{L}_{\text{penalty}} = \mathbb{E}_{X}\left[ \sum_{l=1}^{k} \mathbb{I}[q^\ast_{i_l}(x) \neq \hat{q}_{i_l}(x)]  \right],
\label{eqn:penalty}
\end{equation}
where \( \hat{q}_{i}(x)\), \( {q}^\ast_{i}(x)\) denote the predicted and ground-truth responses for query \(q_{i}\), respectively, and \( \mathbb{I}[\cdot] \) is the indicator function, which equals 1 if the condition inside the brackets is true and 0 otherwise. The overall loss function is:
\begin{equation}
    \mathcal{L}_{\text{IUAV-IP}} = \mathcal{L}_{\text{V-IP}} + \lambda \cdot \mathcal{L}_{\text{penalty}},
    \label{eqn:penalty}
\end{equation}
where \(\lambda\) is a hyperparameter we tune using a validation set.

\vspace{.5em}
\noindent \textbf{Hybrid Stopping Criterion:}
\label{sec:HSP}
Since the V-IP framework selects queries sequentially, a critical design decision is determining when to stop querying. Prior work has utilized threshold- or stability-based criteria, which halt the process once a predefined threshold is reached or when no additional information is gained. While intuitive, this approaches can either terminate too early, missing informative concepts, or continue unnecessarily, resulting in longer explanations without meaningful performance gains. As shown in Table~\ref{tab:iuav_ip_combined}, this often leads to an increased number of queries without corresponding improvements in accuracy.

To address these limitations, we adopt a hybrid confidence-plateau criterion that jointly considers both prediction certainty and stability. Specifically, we monitor the classifier's predicted confidence scores across steps and stop querying when the following condition is met:
\begin{equation}
\left[c_t \geq \tau \right] \wedge  \left[ \frac{1}{w-1} \sum_{j=t-w+1}^{k-1} \left| c_{j+1} - c_j \right| < \delta \right].
\label{equ:stop}
\end{equation}
Here  \( c_t = \max f_\eta(\mathcal{H}_{1:k}^\pi(\cdot)) \) denote the classifier’s confidence at step \( k \),  \( w \) is the window length and \( \delta, \tau \) are the thresholds; $\wedge$ denotes the logical AND operator.

The stopping criterion is evaluated as a conjunction of both conditions.
Specifically, the querying process halts only when both the confidence exceeds the predefined threshold (\(c_t \geq \tau\)) and the variation in confidence within the recent window (\(w\)) is smaller than \(\delta\).
This ensures that the model stops only when it is simultaneously confident and stable, thereby preventing both premature termination and redundant querying.

\subsubsection{Query-Wise Uncertainty Quantification}
\label{sec:uncertainityQuant}
Uncertainty estimation is a critical component in safety-sensitive domains like medical imaging. Prior work has proposed various approaches for quantifying uncertainty in deep neural networks~\cite{gal2016dropout,kendall2017uncertainties,wimmer2023quantifying,gawlikowski2023survey}. In this  study, we adopted two widely used techniques to quantify uncertainty in the outputs of the CP \( q_\theta \).

\vspace{.5em}
\noindent \textbf{Entropy-based Uncertainty:} 
Given an input image \( x \) and a query \( q_m \), the concept predictor outputs a predictive distribution \( \Phi^{(m)} \in \mathbb{P}(\mathcal{C}) \), where each element \( \phi_k^{(m)} = P(C = c_k \mid x, \theta) \), and \( k \in \{-1, +1\} \) due to binary outputs. The entropy of this distribution captures uncertainty as:
\begin{equation}
    \mathcal{U}_{q_m}^{\mathcal{H}}(x) = -\sum_{k \in \{-1, +1\}} \phi_k^{(m)} \log_2(\phi_k^{(m)}).
    \label{eqn:entropy_uncertainity}
\end{equation}

\vspace{.5em}
\noindent \textbf{Monte Carlo Dropout-based Uncertainty:} 
Following Sale et al.~\cite{sale2024label}, we also estimate uncertainty using a second-order distribution (i.e., \(\mathbb{P}(\mathbb{P}(\mathcal{C})) \)) over predicted probabilities.We use MC dropout as a computationally efficient approximation to the predictive distribution~\cite{gal2016dropout}. In finite-data settings with annotation variability, aleatoric and epistemic uncertainties are entangled, making predictive variance a practical proxy for overall uncertainty and enabling uncertainty-aware query selection~\cite{hullermeier2021aleatoric}. Let \( \Pi^{(m)} \in \mathbb{P}(\mathbb{P}(\mathcal{C})) \) be the distribution over predictive probabilities obtained via multiple stochastic forward passes. The total uncertainty is \cite{sale2024label}:
\begin{equation}
    \mathrm{TU}(q_m) = \underbrace{\mathbb{E}\left[ \Phi^{(m)} (1 - \Phi^{(m)}) \right]}_{\text{Aleatoric}} + \underbrace{\mathrm{Var}\left( \Phi^{(m)} \right)}_{\text{Epistemic}}.
\end{equation}
where \(_{\Phi^{(m)}\sim \Pi_m}\). 
This can be approximated with finite number of MC samples \( \mathcal{S} \):
\begin{equation}
    \mathcal{U}_{q_m}^{\text{MC}}(x) = \frac{1}{|\mathcal{S}|} \sum_{s \in \mathcal{S}} \Phi_m^{(s)}(1 - \Phi_m^{(s)}) + \frac{1}{|\mathcal{S}|} \sum_{s \in \mathcal{S}} \left( \Phi_m^{(s)} - \bar{\Phi}_m \right)^2.
    \label{eqn:Monte Carlo_uncertainity}
\end{equation}

\subsection{Experimental Results}
\subsubsection{Quantative Evaluation}
The performance of the proposed EUAV-IP and IUAV-IP models across five datasets is summarised in Table~\ref{tab:iuav_ip_metrics}. All methods were run 10 times with different random seeds and train-validation-test (70-10-20, except for WBCatt, which follows the official split provided by the dataset curators) splits; mean and standard deviation are reported.
We compare our proposed methods against a range of baselines, including a BlackBox classifier (ResNet101) and a decision tree-based interpretable model (CBM-DT~\cite{quinlan1986induction}), where a decision tree is used to classify the outputs from a CP.
We also include comparisons with the standard CBM~\cite{koh2020concept} and its recent uncertainty-aware variants, ProbCBM~\cite{kim2023probabilistic} and evi-CEM~\cite{gao2024evidential}.
These two models are the only CBM-based approaches that explicitly incorporate uncertainty estimation, making them the most relevant references for our study.
Other CBM and V-IP extensions such as LF-CBM~\cite{oikarinen2023label}, LaBo~\cite{yang2023language}, Bootstrapping V-IP \cite{chattopadhyay2024bootstrapping}, and Coherent-CBM~\cite{patricio2023coherent} focus primarily on improving concept curation, semantic alignment, or spatial coherence of concept activations rather than uncertainty handling, and thus are complementary but not directly comparable.

Among V-IP-based methods, we evaluate the original V-IP~\cite{chattopadhyayvariational}, two explicit uncertainty-aware variants, EUAV-IP (Entropy) and EUAV-IP (Monte Carlo), two implicit uncertainty-aware variants, IUAV-IP (Entropy) and IUAV-IP (Monte Carlo), and an oracle-based version, UAV-IP (Oracle), which uses ground-truth concepts to mask incorrect answers.
While the oracle provides an upper bound on performance, it is not feasible in real-world scenarios where concept labels are unavailable at inference time.
Our comparisons therefore focus on methods that directly address uncertainty modeling, alongside standard interpretable and black-box baselines.

Among the four variants of our proposed method, the IUAV-IP (Monte Carlo) variant consistently achieves high accuracy across all five datasets while maintaining shorter query lengths. This demonstrates that implicit uncertainty integration is more effective than explicit integration—partly because explicit methods rely on thresholds, whereas implicit methods can leverage uncertainty without such constraints. Additionally, the results indicate that Monte Carlo–based uncertainty estimation is more effective than entropy-based estimation. Further discussion of these findings is provided in Section~\ref{sec:uncertainity_effectiveneess}. The proposed variants significantly outperform the original V-IP~\cite{chattopadhyayvariational} method, demonstrating the effectiveness of incorporating uncertainty. On average, IUAV-IP (Monte Carlo) achieves around a 5\% improvement in accuracy while using 20\% fewer queries.

IUAV-IP (Monte Carlo) achieves near-oracle performance on most datasets, despite not having access to ground-truth concepts. The only exception is the OAI dataset, where limited CP (see Table \ref{tab:auc_scores}) quality hinders accurate uncertainty estimation. While the BlackBox model achieves high accuracy, it lacks transparency and interpretability, providing no explanation for its decisions.

Furthermore, IUAV-IP (Monte Carlo) outperforms CBM~\cite{koh2020concept} and its uncertainty-aware variants (ProbCBM~\cite{kim2023probabilistic} and evi-CEM~\cite{gao2024evidential}) across all datasets, while also reducing the number of queries required to reach a decision.\footnote{While models like ProbCBM and evi-CEM are designed to incorporate human feedback or interventions to adjust concept responses, we did not utilise such interventions in our experiments to ensure a fair comparison across methods.} For example, on the Derm7pt dataset, CBM-based approaches require all 38 concepts for each prediction, whereas IUAV-IP, on average, uses only 23 informative queries. This selective querying reduces explanation length and better mirrors clinical reasoning, where decisions are often based on a few key observations rather than exhaustive assessments. Thus, IUAV-IP provides more concise and intuitive explanations that are easier for domain experts to interpret while also being more accurate. 

\subsubsection{Reliability Evaluation} 
To assess the robustness of query selection, we use two additional metrics: Step Error Rate (SER) and First Wrong Query (FWQ). 
The FWQ asked metric tracks when a model incorporates a concept with a wrong answer. Ideally, models that utilize uncertainty will be able to determine the likelihood of a concept being wrong and postpone relying on it, thus having a higher FWQ. As shown in Table \ref{tab:iuav_ip_metrics}, in all four datasets, IUAV-IP (Monte Carlo) utilizes uncertain queries much later than V-IP, demonstrating the effectiveness of conditioning the querier with uncertainty values and the proposed loss function. 

\begin{figure}
    \centering\includegraphics[width=\linewidth]{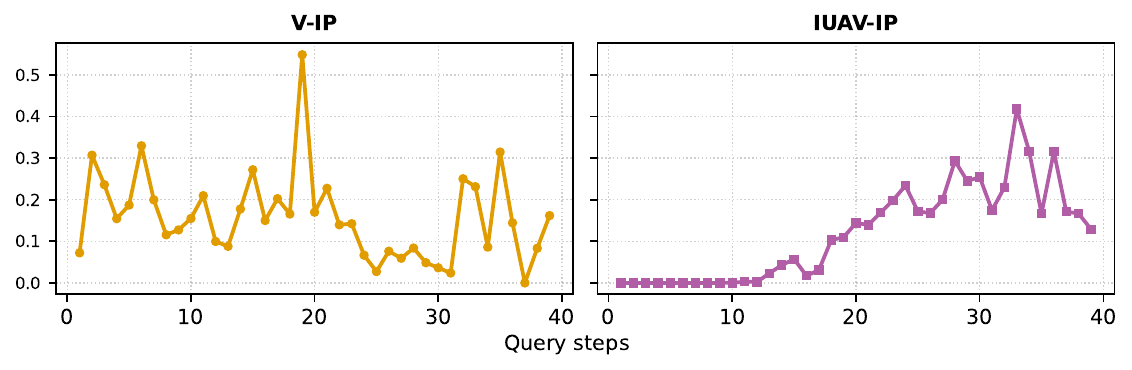} 
    \caption{
    \textbf{Step-wise Error Rate (SER) comparison across models on the Derm7pt dataset.}
    IUAV-IP (right) maintains a low SER in early steps, with errors rising only in later queries-indicating more informed and cautious reasoning. In contrast, V-IP shows earlier spikes in SER, reflecting unstable or premature query decisions.
    }
    \label{fig:ser_plot}
\end{figure}

Figure~\ref{fig:ser_plot} shows the SER per query index across all test samples in the Derm7pt dataset. SER measures how often the model selects an incorrect concept at each step. A lower SER indicates more reliable queries are being selected. However, given that our method does not improve the accuracy of the CP, the total number of errors will stay the same between methods, however by prioritizing correct queries at the start and stopping early, we can assure that more reliable queries are used for classification.
As shown in the right plot, IUAV-IP exhibits a steadily increasing SER curve that remains relatively low and stable until around step 16-18. This behavior shows that IUAV-IP prioritizes the most certain (and thus reliable) queries early in the decision process.  This corresponds with our quantitative results in Table~\ref{tab:iuav_ip_metrics}, where IUAV-IP achieves high accuracy using fewer queries.
In contrast, V-IP shows noisy SER trends, especially in the early stages, reflecting an inability to determine query reliability. These results indicate that the utility of uncertainty-aware querying scales with the difficulty of the task: it provides robust safety checks on easier datasets, while driving significant performance gains in more challenging, clinically realistic settings.

\subsubsection{Qualitative Analysis}
\begin{figure*}[t!]
\centering
\includegraphics[width=.9\textwidth]{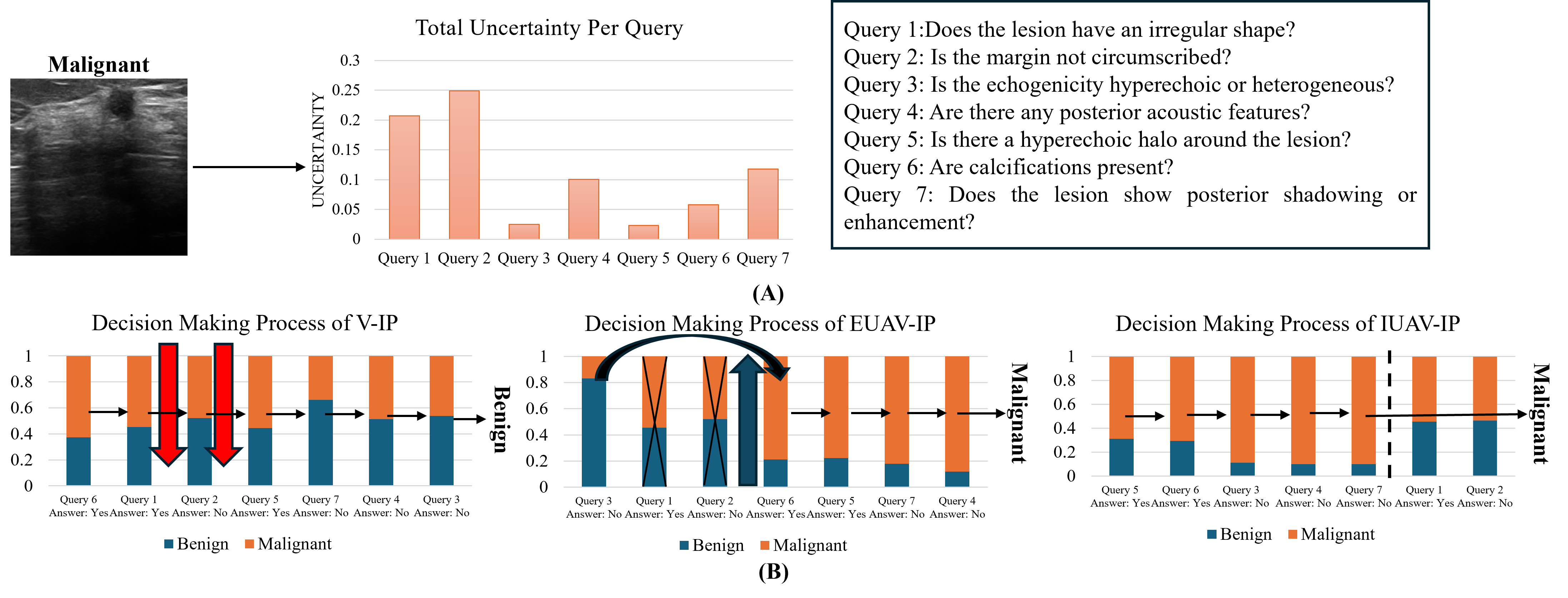}
\caption{ 
\textbf{Qualitative comparison of query-based reasoning across models for a malignant ultrasound image from the BrEaST dataset.}
\textbf{(A)} Total uncertainty per query computed using Monte Carlo Dropout (Eq.~\ref{eqn:Monte Carlo_uncertainity}). Queries 1 and 2 exhibit high uncertainty, while Queries 5 and 6 are more confident and reliable.
\textbf{(B)} Decision-making trajectories for three models: V-IP, EUAV-IP, and IUAV-IP.
V-IP selects highly uncertain and incorrect queries (red arrows), leading to a misclassification (\textit{Benign}), whereas EUAV-IP effectively skips these uncertain queries.
IUAV-IP dynamically balances informativeness and uncertainty, achieving the correct prediction (\textit{Malignant}) with stable confidence without explicit masking.
}
\label{fig:qualitative_breast}
\end{figure*}

We also present a qualitative case study from the BrEaST dataset in Figure~\ref{fig:qualitative_breast}. The input image, labeled as \textit{Malignant}, was evaluated using the V-IP, EUAV-IP, and our proposed IUAV-IP models. First, we computed the total uncertainty per query using the Monte Carlo-based uncertainty formulation in Eq.~\ref{eqn:Monte Carlo_uncertainity}, which is shown in panel (A).

\vspace{.5em}
\noindent \textbf{V-IP Behavior.} Since the original V-IP framework does not consider uncertainty, it selects the next query solely based on informativeness. As shown in panel (B-left), it initially selects an informative concept (\textit{Query 6: Answer = Yes}) and slightly improves its confidence. However, in subsequent steps, it selects high-uncertainty queries (e.g., \textit{Query 1, 2}) whose answers are incorrect. These errors significantly reduce its belief in the correct class. Consequently, the model's confidence fluctuates and ultimately results in a misclassification, labeling a malignant case as benign.
This behavior highlights that V-IP learns a global querying policy from the training distribution, where informativeness is measured in expectation over all samples. While this formulation optimizes dataset-level information gain, it fails to adapt to instance-specific uncertainty at test time. For example, although V-IP learns during training that Queries 1 and 2 are generally informative,they become unreliable for the shown test case, causing unstable reasoning. This inability to adapt to input-dependent uncertainty leads to inconsistent decision trajectories when a sample deviates from the training uncertainty profile.

\vspace{.5em}
\noindent \textbf{EUAV-IP Behavior.} In contrast, EUAV-IP explicitly masks high-uncertainty queries and avoids using them. As seen in panel (B-middle), it skips uncertain queries (e.g., \textit{Query 1, 2}) and focuses on more reliable ones such as \textit{Query 6} and \textit{Query 5}. After just five steps, the model confidently predicts \textit{Malignant} with over 90\% confidence.

\vspace{.5em}
\noindent \textbf{IUAV-IP Behavior.} IUAV-IP achieves a similar outcome by utilizing concept predictions together with their uncertainty estimates. Panel (B-right) illustrates this reasoning process: IUAV-IP begins with a highly certain and relevant concept (\textit{Query 5}), then selects another informative and confident query (\textit{Query 6}). As more queries are selected, the model builds confidence gradually and stabilizes above the decision threshold by step 5, leading to an accurate prediction \textit{Malignant}.
This example demonstrates how IUAV-IP, unlike the original V-IP, dynamically adjusts its querying behavior to each input by weighting both informativeness and uncertainty. While V-IP captures global informativeness from the training distribution, IUAV-IP introduces a mechanism for instance-specific adaptation, allowing the model to reason robustly even when concept uncertainties differ from the learned training prior.

\subsubsection{Ablation Study}
\label{sec:AS}
To assess the impact of different uncertainty estimation methods and our two novel contributions: (1) a loss penalty that prioritizes more certain queries, and (2) a novel stopping criterion. We analyze the model's performance with different configurations in an ablation study.

As shown in Table \ref{tab:iuav_ip_metrics}, the second-order Monte Carlo-based uncertainty estimates yielded slightly higher AUCs compared to first-order entropy-based estimates. This improvement may be attributed to the Monte Carlo method's ability to capture both epistemic (model) and aleatoric (data) uncertainties, whereas entropy primarily reflects epistemic uncertainty. Accurately estimating both types of uncertainty is crucial in medical imaging applications to enhance model reliability and trustworthiness.

\begin{figure}[ht]
    \centering
    \includegraphics[width=0.6\linewidth]{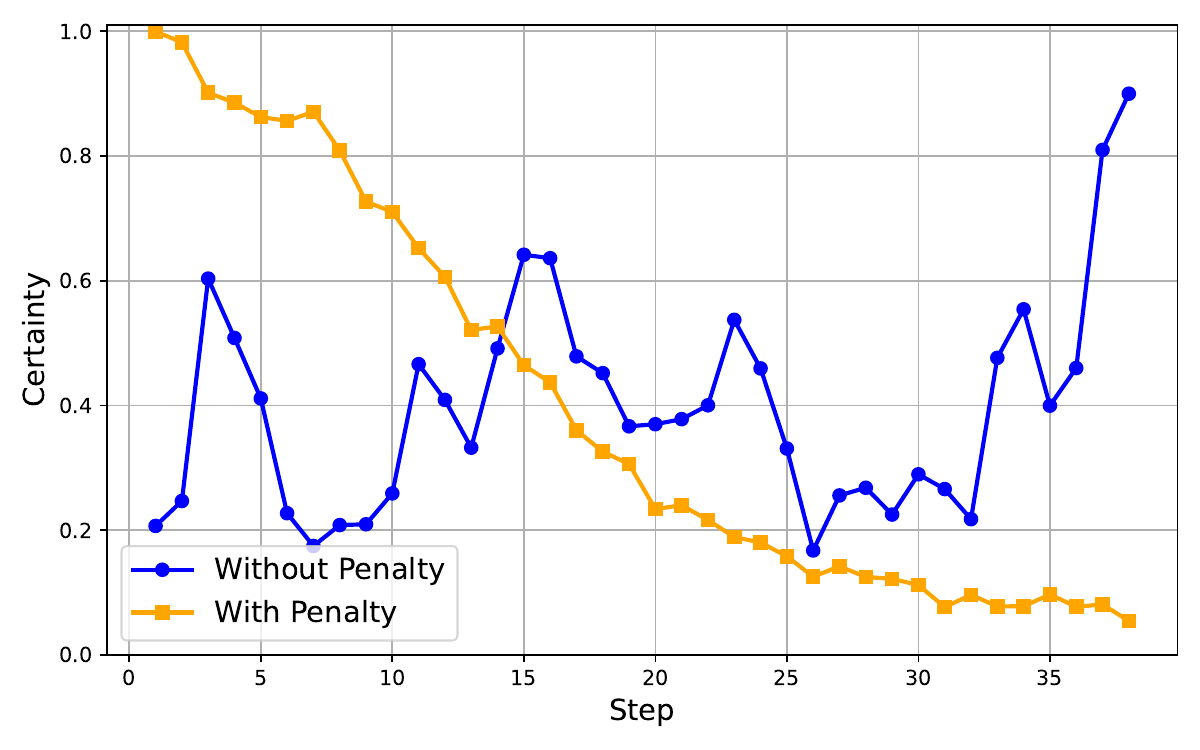}
    \caption{Query certainty trends over steps in Derm7pt dataset. IUAV-IP (Monte Carlo) with penalty (orange) begins with high-certainty queries and gradually transitions to more uncertain ones, while the version without penalty (blue) exhibits noisier and less structured query selection.}
    \label{fig:penalty_comparison}
\end{figure}

Table \ref{tab:iuav_ip_combined} presents the effectiveness of incorporating the penalty term (P) and the hybrid stopping criterion (HS) in the IUAV-IP (Monte Carlo) model. The results indicate that the IUAV-IP model, when augmented with both the penalty term and the hybrid stopping criterion, achieves higher performance compared to configurations where these components are omitted. Notably, the inclusion of the penalty term contributes more significantly to the overall performance enhancement.
Figure~\ref{fig:penalty_comparison} further substantiates these findings. In the Derm7pt dataset (with similar trends observed in other datasets), the IUAV-IP (Monte Carlo) model with the penalty term initiates with confident queries and gradually transitions to more uncertain ones. Conversely, the variant without the penalty term displays erratic certainty patterns, indicating less effective prioritization. Collectively, these quantitative and qualitative insights validate our hypothesis that effective uncertainty-aware reasoning necessitates both input-awareness and penalty-based supervision.

The results in Table~\ref{tab:iuav_ip_combined} consistently demonstrate that the hybrid stopping approach reduces the average number of queries and improves accuracy and reliability.

\subsubsection{Calibration Performance}
To further assess model reliability, we evaluated the calibration quality of the classifier using the \textit{Expected Calibration Error (ECE)} metric. 
As shown in Table~\ref{tab:ece_results}, our proposed IUAV-IP framework achieves consistently lower ECE values compared to the baseline V-IP across multiple datasets.
This demonstrates that uncertainty-aware querying leads to better-calibrated confidence estimates, improving the trustworthiness and interpretability of model predictions.
The oracle-based UAV-IP (Oracle) provides a reference upper bound for calibration quality. 

\begin{figure}[t!]
    \centering
    \includegraphics[width=\linewidth]{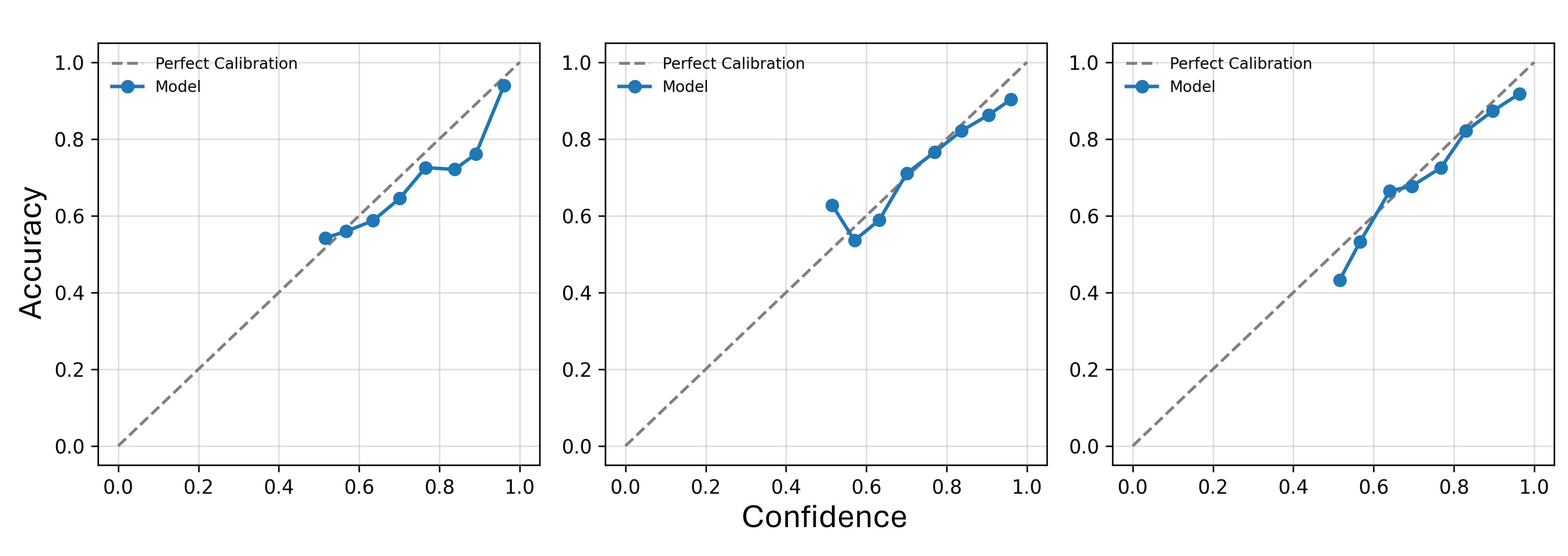}
    \caption{
    \textbf{Reliability diagrams comparing calibration behaviour across models on the Derm7pt dataset.}
    The dashed diagonal represents perfect calibration, where predicted confidence equals observed accuracy.
    The baseline V-IP (left) exhibits mild overconfidence.
    The proposed IUAV-IP (Monte Carlo) (middle) and oracle UAV-IP (right) achieve more consistent alignment with the ideal calibration curve.
    }
    \label{fig:calibration_reliability}
\end{figure}

To visualize the calibration quality of model predictions, Figure~\ref{fig:calibration_reliability} presents reliability diagrams that relate predicted confidence to empirical accuracy across different methods.
The diagonal line indicates perfect calibration, where confidence values correspond exactly to the likelihood of being correct. As shown, the baseline V-IP model (left) exhibits slight overconfidence, along with some inconsistency across confidence intervals. In contrast, the IUAV-IP (Monte Carlo) variant (middle) demonstrates a closer and smoother fit to the ideal diagonal, suggesting that its uncertainty-aware query mechanism yields more reliable confidence estimates.
Finally, the oracle UAV-IP (right) exhibits near-perfect calibration, providing an upper bound for achievable reliability.

Overall, these results indicate that integrating uncertainty modeling within the V-IP framework not only improves classification accuracy and query efficiency but also enhances confidence calibration, which is a critical aspect for safe and interpretable decision-making in medical AI systems.

\begin{table}

\centering
\caption{Expected Calibration Error (ECE) comparison across datasets. 
\textbf{Bold} indicates the best non-oracle result.}
\label{tab:ece_results}
\scriptsize
\renewcommand{\arraystretch}{1.1}
\begin{tabular}{|c|l|c|}
\hline
\textbf{Dataset} & \textbf{Model} & \textbf{ECE}~$\downarrow$ \\
\hline
\multirow{3}{*}{Derm7pt} 
& V-IP & 0.0615 ± 0.0005 \\
& IUAV-IP (MC) & \textbf{0.0374 ± 0.0006} \\
\rowcolor{oraclegray}
& UAV-IP (Oracle) & 0.0293 ± 0.0007 \\
\hline
\multirow{3}{*}{BrEaST} 
& V-IP & 0.0574 ± 0.0005 \\
& IUAV-IP (MC) & \textbf{0.0545 ± 0.0005} \\
\rowcolor{oraclegray}
& UAV-IP (Oracle) & 0.0267 ± 0.0003 \\
\hline
\multirow{3}{*}{OAI} 
& V-IP & 0.1051 ± 0.0009 \\
& IUAV-IP (MC) & \textbf{0.0761 ± 0.0008} \\
\rowcolor{oraclegray}
& UAV-IP (Oracle) & 0.0752 ± 0.0009 \\
\hline
\end{tabular}
\end{table}

\begin{table}[t]
\centering
\caption{
Effectiveness of uncertainty estimates in classifying query-answer correctness, measured using AUC (discrimination ability) and ECE (calibration error). Monte Carlo and Entropy represent two uncertainty estimation strategies. ConceptAUC denotes the discrimination performance of the concept predictor model.
}
\label{tab:auc_scores}
\setlength{\tabcolsep}{4pt}
\begin{tabular}{lccccc}
\toprule
& \multicolumn{2}{c}{\textbf{AUC (\%)}} 
& \multicolumn{2}{c}{\textbf{ECE}} 
& \textbf{ConceptAUC (\%)} \\
\cmidrule(lr){2-3} \cmidrule(lr){4-5}
\textbf{Dataset} 
& \textbf{MC} & \textbf{Entropy} 
& \textbf{MC} & \textbf{Entropy}
& \\
\midrule
PH2      & 95.75 & 91.19 & 0.05$\pm$0.02 & 0.41$\pm$0.15 & 99.48 \\
Derm7pt  & 84.59 & 74.64 & 0.07$\pm$0.02 & 0.42$\pm$0.15 & 70.71 \\
BrEaST   & 91.00 & 71.30 & 0.10$\pm$0.03 & 0.45$\pm$0.09 & 95.94 \\
WBCatt   & 90.39 & 90.19 & 0.05$\pm$0.02 & 0.37$\pm$0.16 & 95.09 \\
OAI      & 79.84 & 77.00 & 0.12$\pm$0.03 & 0.46$\pm$0.17 & 73.68 \\
\bottomrule
\end{tabular}
\end{table}

\begin{table}[ht]
\centering
\caption{Comparison of IUAV-IP across combinations of Penalty (P) and Hybrid Stopping (HS). 
\textbf{Bold} indicates best performance per dataset.}
\label{tab:iuav_ip_combined}
\renewcommand{\arraystretch}{1.1}
\scriptsize
\begin{tabular}{|l|c|c|c|c|c|}
\hline
\textbf{Dataset} & \textbf{P} & \textbf{HS} & \textbf{Accuracy~$\uparrow$} & \textbf{AUC~$\uparrow$} & \textbf{Queries~$\downarrow$} \\
\hline
\multirow{3}{*}{Derm7pt} 
& \xmark & \xmark & 82.16 ± 2.21 & 89.98 ± 2.14 & 27.10 ± 0.21 \\
& \xmark & \cmark & 84.68 ± 2.50 & 91.62 ± 3.28 & 26.78 ± 0.46 \\
& \cmark & \xmark & 81.33 ± 1.75 & 88.26 ± 1.81 & 26.45 ± 0.70 \\
& \cmark & \cmark & \textbf{90.36 ± 3.88} & \textbf{93.59 ± 4.14} & \textbf{23.52 ± 0.70} \\
\hline
\multirow{3}{*}{BrEaST} 
& \xmark & \xmark & 88.03 ± 2.53 & 94.22 ± 1.43 & 6.89 ± 0.03 \\
& \xmark & \cmark & 89.00 ± 2.65 & 95.70 ± 1.82 & 6.69 ± 0.43 \\
& \cmark & \xmark & 88.12 ± 2.82 & 94.34 ± 1.87 & \textbf{6.45 ± 0.72} \\
& \cmark & \cmark & \textbf{89.60 ± 3.20} & \textbf{95.83 ± 1.18} & 6.96 ± 0.03 \\
\hline
\multirow{3}{*}{OAI} 
& \xmark & \xmark & 53.14 ± 1.57  & 70.19 ± 0.21 & 9.88 ± 0.13 \\
& \xmark & \cmark & 54.12 ± 1.22  & 72.14 ± 0.08 & 9.32 ± 0.68 \\
& \cmark & \xmark & 53.57 ± 0.28 & 70.36 ± 0.23 & 9.81 ± 0.28 \\
& \cmark & \cmark &  \textbf{60.17 ± 0.14} & \textbf{75.14 ± 0.09} &  \textbf{7.15 ± 0.16}  \\
\hline

\end{tabular}
\end{table}

\begin{table}[t]
\centering
\caption{Validation that querying a subset of concepts yields strong performance. \textbf{Bold} indicates the best non-oracle result.}
\label{tab:iuav_ip_subset_validation}
\scriptsize
\renewcommand{\arraystretch}{1.1}
\begin{tabular}{|c|l|c|c|c|}
\hline
\textbf{Dataset} & \textbf{Model} & \textbf{Accuracy}~$\uparrow$ & \textbf{AUC}~$\uparrow$ & \textbf{Queries}~$\downarrow$ \\
\hline
\multirow{4}{*}{Derm7pt} 
& V-IP (NS) & 81.21 ± 1.00 & 87.40 ± 2.17 & 38.00 ± 0.00 \\
& EUAV-IP & 83.49 ± 1.26 & 89.99 ± 1.49 & 23.65 ± 0.52 \\
& IUAV-IP & \textbf{90.36 ± 3.88} & \textbf{93.59 ± 4.14} & \textbf{23.52 ± 0.70} \\
\rowcolor{oraclegray}
& UAV-IP (Ora-NS) & 92.29 ± 1.61 & 97.09 ± 0.57 & 32.56 ± 0.22 \\
\hline
\multirow{4}{*}{BrEaST}  
& V-IP (NS) & 86.49 ± 5.29 & 94.98 ± 2.49 & 7.00 ± 0.00 \\ 
& EUAV-IP & 88.14 ± 3.97 & 95.05 ± 2.29 & \textbf{6.57 ± 0.05} \\
& IUAV-IP & \textbf{89.30 ± 3.20} & \textbf{95.83 ± 1.18} & 6.96 ± 0.03 \\
\rowcolor{oraclegray}
& UAV-IP (Ora-NS) & 89.44 ± 4.13 & 95.57 ± 2.38 & 6.94 ± 0.03 \\
\hline

\multirow{4}{*}{OAI}  
& V-IP (NS) & 52.57 ± 0.36 & 68.80 ± 0.62 & 10.00 ± 0.00  \\
& EUAV-IP &  51.33 ± 0.44 & 65.87 ± 0.65 & 8.16 ± 0.13 \\
& IUAV-IP & \textbf{60.17 ± 0.14} & \textbf{75.14 ± 0.09} &  \textbf{7.15 ± 0.16}  \\
\rowcolor{oraclegray}
& UAV-IP (Ora-NS) &  71.76 ± 0.39  &  88.39 ± 0.70 &  8.14 ± 0.00  \\
\hline
\end{tabular}
\end{table}


\section{Discussion}
\label{sec:discussion}

\subsection{Uncertainty as a Proxy for Reliability}
\label{sec:uncertainity_effectiveneess}
A key hypothesis of this work, which explains the superior performance of our proposed method, is that concept-level uncertainty, derived using either entropy or Monte Carlo Dropout, serves as a reliable heuristic for the correctness of predicted concept values. Table~\ref{tab:auc_scores} presents the average AUC for binary classification of correct vs.\ incorrect concept answers using uncertainty estimates. Both Monte Carlo and entropy estimates achieve strong discrimination of concept correctness. However, the second-order Monte Carlo-based estimates consistently outperform entropy-based ones. For example, in the Derm7pt dataset, Monte Carlo achieves an AUC of 84.59\%, compared to 74.64\% for entropy. 

Importantly, Monte Carlo uncertainty is also substantially better calibrated, as reflected by its lower ECE. Across all datasets, MC-ECE remains consistently small, whereas entropy-based uncertainty exhibits significantly higher calibration error, confirming that Monte Carlo provides both superior discrimination and more trustworthy uncertainty estimates. Furthermore, as shown in Table~\ref{tab:iuav_ip_metrics}, the more reliable the uncertainty estimator, the better the performance.

\vspace{0.5em}
\noindent
\textbf{Choice and generality of uncertainty estimators.}
This work represents the first attempt to incorporate uncertainty estimation within the V-IP framework.
Our objective was not to design new uncertainty measures, but to establish a flexible, estimator-agnostic formulation that allows V-IP to operate with any per-concept uncertainty signal.
In this preliminary study, we used two well-established and computationally efficient estimators, entropy and Monte Carlo dropout, to capture aleatoric and epistemic components, respectively \cite{kendall2017uncertainties,gal2016dropout,wimmer2023quantifying,sale2024label,gawlikowski2023survey}, which are also adopted in medical imaging \cite{jungo2020analyzing}.
On top of that, the proposed EUAV-IP and IUAV-IP frameworks can readily incorporate alternative or domain-specific estimators, such as evidential learning, deep ensembles, or shift-aware uncertainty quantifiers, without altering the overall algorithmic structure.
Exploring these alternatives constitutes a promising direction for future work.

\subsection{Dynamic Subset Selection for Explainability and Uncertainty Handling}
A key idea in the V-IP framework is that basing predictions on a subset of concepts improves explainability, reflected in the evaluation of query length, where shorter is better.

This is particularly relevant to medical imaging for two reasons. First, it mirrors how clinicians make decisions: when reporting a diagnosis, they focus on the most relevant concepts rather than listing all possible ones. Second, by dynamically selecting a subset of concepts for each data point, the model can disregard those that are uncertain and base its decisions solely on the most reliable information available.

While V-IP captures the first behaviour, it lacks the ability to address the second. Our method adds this capability.

To demonstrate that shorter query histories lead to better decisions, we conducted an experiment with the V-IP framework. We removed its stopping criterion, creating two variants: V-IP(NS), which uses all concepts, and UAV-IP(Ora-NS), which uses only correctly predicted concepts. As shown in Table \ref{tab:iuav_ip_subset_validation}, UAV-IP(Ora-NS), using only correctly classified concepts, achieves the best performance. However, this is unrealistic since concept labels are not available at inference. In contrast, our proposed method achieves performance close to UAV-IP(Ora-NS) without using any concept annotations. This highlights the benefits of selecting relevant subsets for decisions and doing so dynamically based on uncertainty.

\section{Conclusion}
In this work, we addressed a critical gap in interpretable AI for medical imaging by introducing uncertainty-aware extensions to the V-IP framework. We propose two models: EUAV-IP, which skips uncertain concepts, and the novel IUAV-IP, which integrates uncertainty estimates into the query selection process to enable more informed, cautious, and clinically aligned decision-making. By learning to prioritise reliable concepts and deferring uncertain ones, the model mirrors the diagnostic reasoning of medical professionals, balancing informativeness with trustworthiness. Through extensive evaluation across diverse medical imaging datasets, we demonstrated that IUAV-IP not only improves predictive performance but also produces shorter, more meaningful explanations. These improvements are not merely technical, they are essential for real-world deployment in clinical settings, where interpretability, reliability, and efficiency are paramount. The ability to delay or avoid uncertain queries without sacrificing accuracy enhances the model’s safety and transparency that are important for responsible use of AI in healthcare. 

While our approach still relies on concept-level supervision during training, it lays the groundwork for future systems that can autonomously generate and reason over clinically relevant concepts using vision-language models. Ultimately, this work moves us closer to AI systems that can collaborate with clinicians, not just as accurate tools, but as interpretable and trustworthy partners in decision-making. On top of that guiding the querier to avoid unreliable queries requires access to concept-level ground truth during training, which may not always be feasible in practice. In future work, we aim to explore more flexible loss functions that can enable effective query selection without requiring explicit concept labels. Additionally, we plan to integrate IUAV-IP with VLMs and LLMs by enabling them to both generate and answer concept-based queries under uncertainty.

\bibliographystyle{unsrt}  
\bibliography{references}  

\end{document}